\documentclass[a4paper, conference]{IEEEtran}
\IEEEoverridecommandlockouts
\usepackage[noadjust]{cite}

\usepackage{amsmath,amssymb,amsfonts}
\usepackage{algorithmic}
\usepackage{graphicx}
\usepackage{textcomp}
\usepackage{xcolor}
\usepackage{float}
\usepackage{array}
\usepackage{tabu}
\usepackage{multirow}
\usepackage{flafter}
\usepackage{tikz}
\usepackage{textcomp}
\usepackage{hyperref}
\usepackage{lipsum}
\usepackage{geometry}
\renewcommand\IEEEkeywordsname{Keywords}
\def\BibTeX{{\rm B\kern-.05em{\sc i\kern-.025em b}\kern-.08em
    T\kern-.1667em\lower.7ex\hbox{E}\kern-.125emX}}

\newcommand\copyrighttext{%
  \footnotesize \textcopyright 2019 IEEE. This paper is under review in "IEEE International Conference on Robotics, Automation, Artificial-Intelligence and Internet-of-Things, 2019". Personal use of this material is permitted.
  Permission from IEEE must be obtained for all other uses, in any current or future
  media, including reprinting/republishing this material for advertising or promotional
  purposes, creating new collective works, for resale or redistribution to servers or
  lists, or reuse of any copyrighted component of this work in other works.}
\newcommand\copyrightnotice{%
\begin{tikzpicture}[remember picture,overlay]
\node[anchor=south,yshift=10pt] at (current page.south) {\fbox{\parbox{\dimexpr\textwidth-\fboxsep-\fboxrule\relax}{\copyrighttext}}};
\end{tikzpicture}%
}

\begin{document}

\title{A CNN-based approach to classify cricket bowlers based on their bowling actions\\}

\author{\IEEEauthorblockN{Md Nafee Al Islam }
\IEEEauthorblockA{\textit{Department of Electrical and} \\
\textit{Electronic Engineering}\\
\textit{Ahsanullah University of Science}\\
\textit{and Technology}\\
\textit{seiumiut@gmail.com}\\
\\
}\vspace*{-.5cm}

\\[-6.0ex]
\and
\IEEEauthorblockN{Tanzil Bin Hassan}
\IEEEauthorblockA{\textit{Department of Electrical and} \\
\textit{Electronic Engineering}\\
\textit{Islamic University of Technology}\\
\textit{tanzilhassan@iut-dhaka.edu}\\
\\
}\vspace*{-.5cm}
\\[-6.0ex]
\and
\IEEEauthorblockN{Siamul Karim Khan}
\IEEEauthorblockA{\textit{Department of Computer Science and} \\ \textit{Engineering}\\
\textit{Bangladesh University of}\\
\textit{Engineering and Technology}\\
\textit{siamulkarim@gmail.com}
\\
}\\[-6.0ex]
\vspace*{-.5cm}

}

\maketitle
\copyrightnotice{}
\begin{abstract}
With the advances in hardware technologies and deep learning techniques, it has become feasible to apply these techniques in diverse fields. Convolutional Neural Network (CNN), an architecture from the field of deep learning, has revolutionized Computer Vision. Sports is one of the avenues in which the use of computer vision is thriving. Cricket is a complex game consisting of different types of shots, bowling actions and many other activities. Every bowler, in a game of cricket, bowls with a different bowling action. We leverage this point to identify different bowlers. In this paper, we have proposed a CNN model to identify eighteen different cricket bowlers based on their bowling actions using transfer learning. Additionally, we have created a completely new dataset containing 8100 images of these eighteen bowlers to train the proposed framework and evaluate its performance. We have used the VGG16 model pre-trained with the ImageNet dataset and added a few layers on top of it to build our model. After trying out different strategies, we found that freezing the weights for the first 14 layers of the network and training the rest of the layers works best. Our approach achieves an overall average accuracy of 93.3\% on the test set and converges to a very low cross-entropy loss.
\newline
\end{abstract}

\section{Introduction}
Artificial intelligence and Machine learning technologies are revolutionizing the way of modern life. Deep learning, a subset of machine learning, is being used widely in the field of image and speech recognition. CNNs are very efficient at detecting different details and patterns on image data. With the advancements in the field of computer vision, deep learning is being increasingly used in sports for various purposes. From helping the match officials in their decision making process to helping the athletes in training on physical aspects - use of artificial intelligence is ubiquitous.  Due to the availability of high amount of data in recent years through television broadcasting and improved camera technologies, research based on computer vision in sports activities is expanding.

Cricket is a major sport in many countries. In recent years, cricket has become a matter of interest for deep learning researchers to carry out research based on various actions in the game. One of the sophisticated activities in cricket is bowling. Every bowler bowls with a different bowling action towards the batsman. Even though, at some points in the delivery, the bowling actions of different bowlers may look similar, the overall bowling action is quite unique. Therefore, a system can be developed to detect these unique bowling actions to identify a bowler from an image and video clip. Such a system can be of great use to the broadcasters who have to keep track of the bowlers throughout the whole match.

In this work, we have proposed a method for identifying the bowler from bowling action images. The classifier was built using transfer learning  \cite{torrey2010transfer}. Transfer learning is a method where we use a pre-trained model and modify it to create a separate model. Here, we have employed a famous pre-trained model named VGG16 \cite{simonyan2014very} to build our classifier. We have removed the final layer of the model and modified it by adding three more dense layers and an output layer to build our classifier. We have found that, freezing the weights of the first 14 layers and training the remaining layers provides the best accuracy for the proposed architecture. Also, there is no pre-existing dataset containing cricket bowling action images. So, to train and evaluate the performance of our model, we have built our own dataset and named it “BolwersNet”. The dataset contains 8100 images of 18 different cricket bowlers belonging to seven different cricket playing nations. 

Our proposed system can assist broadcasters, scorers and the team management in many ways. People engaged in broadcasting have to keep track of the bowlers manually. Our system can be used to count the balls bowled by a particular bowler and update the name of the bowler at the beginning of each over. Also, the system can be utilized to gather individual highlights of a particular bowler which can be helpful for match analysis and coaching.

Previously, no other research work has built a classifier to detect bowling actions of cricket bowlers. Also, CNN based models are the latest and most efficient ways for image classification. Thus, our approach to build a CNN based model to classify cricket bowlers based on their bowling actions can be noted as a novel approach in line with the state-of-the art techniques.

\section{Related Works}

Several works utilizing computer vision has been done in the domain of cricket activity recognition. Dixit et al. \cite{dixitdeep} compared three different CNN architectures in terms of ball-by-ball cricket video classification. They used a pre-trained VGG16 CNN architecture for transfer learning to classify each ball into several outcomes. Batra et al. \cite{batra2014implementation} proposed an automated multi-dimensional visual system which detected no-balls bowled in a cricket match. Hari et al. \cite{hari2014event} used intensity projection profile of the umpires to extract the events in cricket match highlights.
Chowdhury et al. \cite{chowdhury2016application} proposed a method to detect foot overstep no-ball using computer vision techniques. Lazarescu et al. \cite{lazarescu1999classifying} classified cricket shots using camera motion parameters. Karmaker et al. \cite{karmaker2015cricket} used batsman motion vector to detect cricket shots. Semwal et al. \cite{semwal2018cricket} used saliency and optical flow to bring out static and dynamic cues from cricket videos and then used CNNs on these cues to extract feature representations. They finally used a Support Vector Machine (SVM) \cite{burges1998tutorial} on these feature representations to classify cricket shots.

Many of the existing works have used image classification techniques for sports activity recognition. It is observed that Convolutional Neural Network based approaches provide the most accurate detections. There have been a few explorations made by the researchers in the domain of Cricket activity recognition. But none of them have worked on recognizing bowling actions of the bowlers. The outcome of our research can be a baseline for the future research works on bowling action recognition.

\section{Basics of CNN}

\begin{figure}[ht]
  \includegraphics[width=\linewidth]{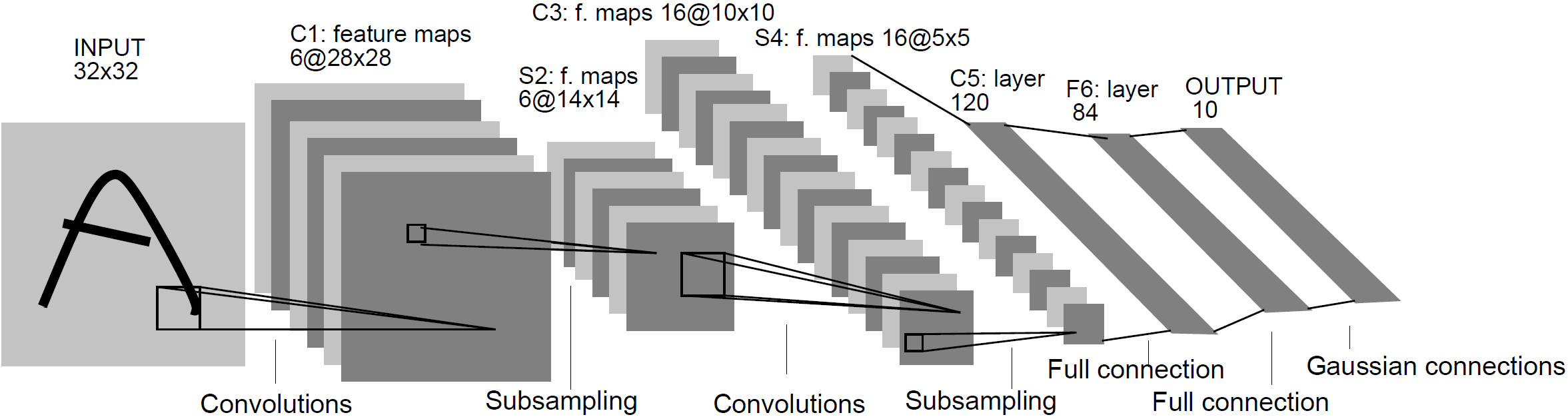}
  \vspace{-10pt}
  \caption{Architecture of LeNet 5. Each plane is a feature map, i.e. a set of units whose weights are constrained to be identical}
  \label{fig:lenet5}
\end{figure}

Convolutional neural network (CNN) \cite{tut1convnet, le2015tutorial} is a deep learning architecture that has been massively fruitful for image classification \cite{krizhevsky2012imagenet}. Yann LeCun et al. \cite{lecun1989} first introduced the idea of a convolutional neural network that can be trained through backpropagation. The CNN architecture became popular among deep learning researchers through the introduction of LeNet-5 in \cite{lecun98} which showed exceptional performance for handwritten character recognition.
The architecture of LeNet-5 is shown in Fig. \ref{fig:lenet5}.

A CNN is constructed with three types of key layers: Convolutional layers, Pooling layers and Fully-connected layers.
The convolutional layer is built with a set of filters or kernels which are convolved over the actual image to extract features from the image in a feature map. At first, the filters or kernels are placed on a particular position on the image. Then element-wise multiplication followed by summation is carried out to create a single pixel value in the feature map. The filter slides over the whole image and performs the same task to create a complete feature map. Equation \ref{eq:1} demonstrates this convolution operation mathematically.
\begin{equation}
 f(x)*g(x) = \int\limits_{-\infty}^{\infty} f(\tau)g(t-\tau) d\tau \label{eq:1}
\end{equation}
On top of the feature maps, a nonlinear activation function is applied to increase nonlinearity \cite{kuo2016understanding}. The most commonly used nonlinear activation functions are Rectified Linear Units (ReLU) \cite{ nair2010rectified}, sigmoid, tanh and softmax (usually used for output layer only).  The initial convolutional layers mostly extract low-level features like vertical and horizontal lines, curves etc. The deeper convolutional layers are able to extract higher-level features like a hand, human body shapes etc.
Usually, a pooling layer is placed after a convolutional layer to reduce the spatial size of the feature map and lessen the computational costs \cite{scherer2010evaluation}. Most commonly practiced pooling methods are max pooling and average pooling \cite{boureau2010theoretical}.
After the convolution and pooling operations, the pixel values are then flattened and reshaped to a single column vector and then fed into an Artificial Neural Network (ANN).  In this ANN, there are one or more fully connected layers where every neuron from the current layer is connected to every neuron of the previous layer. At the end of the fully connected layers, there is an output layer which has number of nodes equal to the number of classes.

\section{Proposed model for classifying bowler}
We propose a CNN-based approach to classify cricket bowlers based on their bowling actions using transfer learning \cite{torrey2010transfer}. Transfer learning is a learning approach where you can take a pre-trained model and replace its output layer with a layer that has the number of nodes you need for your classification. It is an amazing tool to overcome the limitations of small dataset and less-advanced hardware.  The pre-trained model is usually trained on a huge dataset and is already able to detect high level features. Some of the popular pre-trained transfer learning models are VGG16, VGG19, InceptionV3, MobileNet etc. Transfer learning saves a lot of computational expenses, training time and also it allows the model to achieve good accuracy with a minimal amount of data. 

\subsection{VGG16}
\begin{figure*}[ht]
  \centering
  \includegraphics[width=17cm, height=6.1cm]{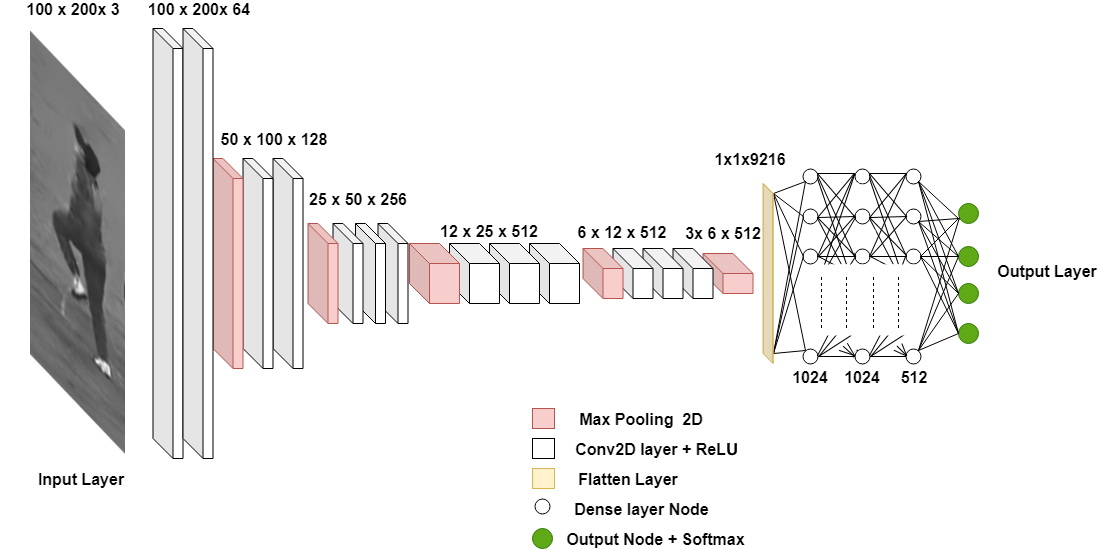}
  \setlength{\belowcaptionskip}{-10pt}
  \caption{Architecture of the proposed Convolutional Neural Network.}
  \label{fig:nafeenet}
\end{figure*}

VGG16 is a famous CNN model which was proposed by K. Simonyan and A. Zisserman \cite{simonyan2014very}. The model has been trained with ImageNet dataset \cite{imagenet_cvpr09} which contains over 14 million images belonging to 1000 different classes. The model performed with a 92.7\% test accuracy on ImageNet data. We used a pre-trained VGG16 model, removed its final layer and added three more dense layers and an output layer for our classification. We kept the weights of first 14 layers unchanged and trained the remaining layers with our dataset. Fig. \ref{fig:nafeenet} shows the architecture we have used for this classification. 

\subsection{Data Collection}
\begin{figure}[ht]
  \centering
  \includegraphics[width=8cm, height=8cm]{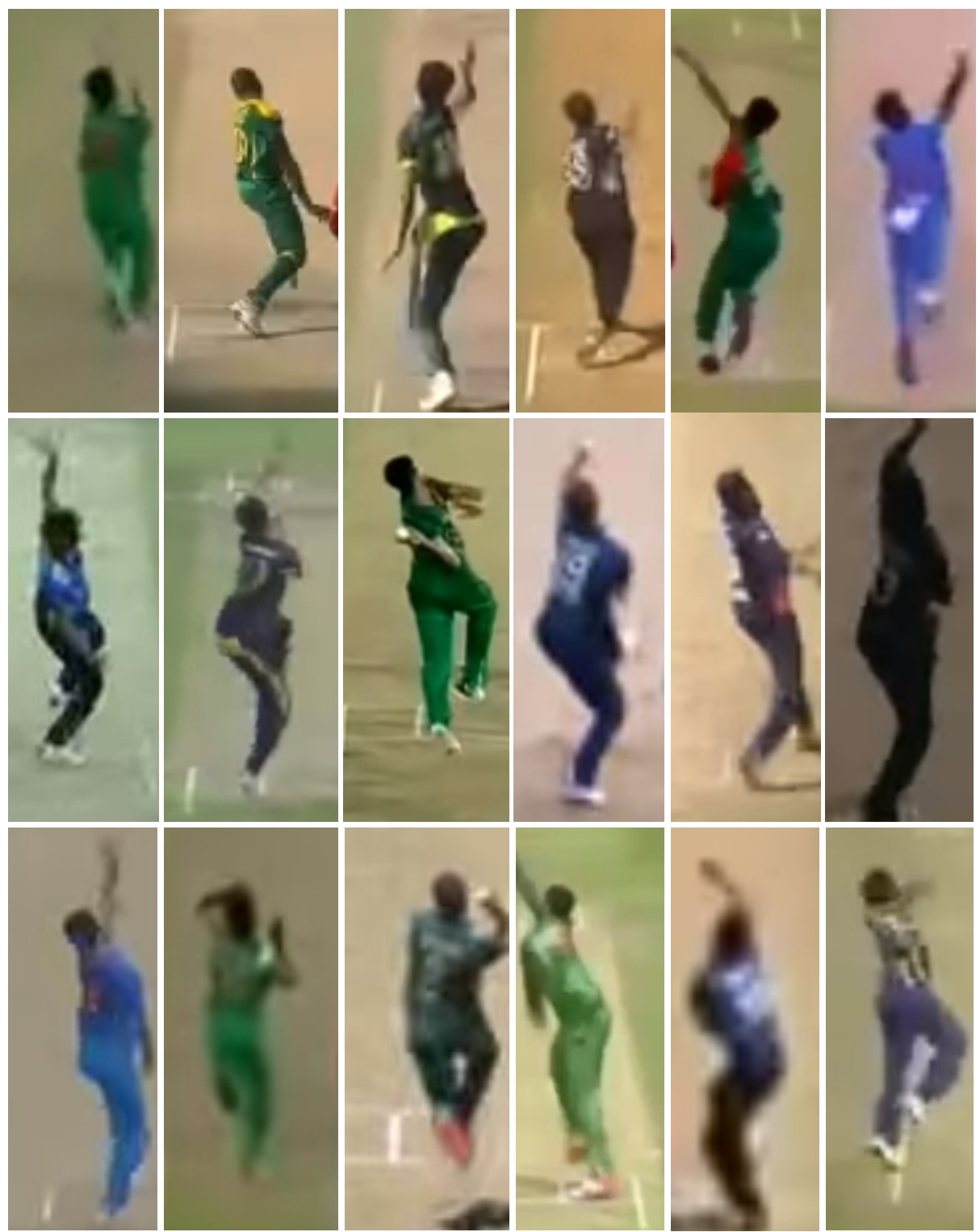}
  \setlength{\belowcaptionskip}{-10pt}
  \vspace{-5pt}
  \caption{A subset of the dataset collected for training and evaluating the model.}
  \vspace{-5pt}
  \label{fig:datasub}
\end{figure}

As we could not find any pre-existing image dataset on bowling action recognition, we had to construct our own dataset –- “"BowlersNet"” for training and testing our model. The dataset contains 8100 images belonging to 18 different bowlers from seven different cricket playing nations. There are 450 images for each of the classes. The images are taken from public videos using snipping and cropping tools by human agent. Images of the bowlers are taken at several distinguishing key points of their bowling actions. From the dataset, 6480 images (360 images of each class) were used for training the model. The remaining 1620 images (90 images of each class) were used as a validation set. Apart from that, we have created an additional test dataset containing 540 images (30 images of each class) for testing purpose which was kept apart from the training and validation process.  Fig. \ref{fig:datasub} shows a subset of our dataset.

\subsection{Data Preprocessing and Augmentation}
Data augmentation is a very effective process to moderate the negative effects of having a small dataset. It helps to avoid overfitting by artificially expanding the volume of the training dataset. This process takes an existing image and performs operations like zooming, shifting, scaling etc. on it to create multiple images. For our training process, we have used a width shift range of 0.3, height shift range of 0.3 and zoom range of 0.2 to augment the data. For our model, all the training data has to be of the same shape. So, we resized all the data to $100 \times 200$ before feeding it to the model as input for training.

\subsection{Training the model}
\begin{figure}
  \centering
  \includegraphics[clip=true, width=7.5cm, height=6.5cm]{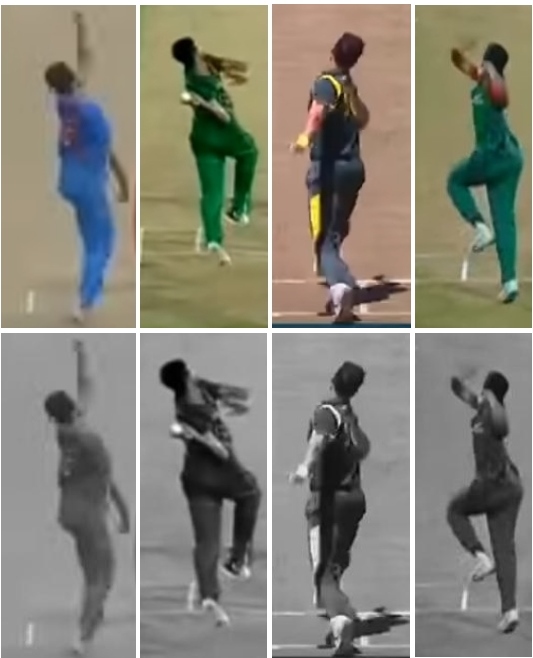}
  \vspace{-5pt}
  \caption{Conversion of some RGB images to grayscale images used for training}
  \vspace{-5pt}
  \label{fig:gray}
\end{figure}

The model was trained with the training set we prepared. After trying out different combinations through trial-and-error process and tuning different parameters, 3 additional dense layers (two having 1024 nodes and one having 512 nodes) and an output layer were added in place of the final layer of the VGG16 model to achieve a good accuracy.  The weights of the first 14 layers of the overall model was kept unchanged and the remaining layers of the model were trained. Softmax activation was used in the final layer of our network.To avoid overfitting, a dropout of 10\% was added to each of the dense layers. To make sure that the model does not overfit on the jersey colors of the bowlers, initially the all training images were converted to grayscale images with a single color channel. And as the VGG16 pre-trained architecture expects images having 3 color channels, the grayscale images were converted to 3 channel grayscale and then fed to the model for training. Fig. \ref{fig:gray} shows some of the RGB images and corresponding converted grayscale images used for training. Different optimizers were tested for the model to reduce the cross entropy to a minimum possible value, and RMSProp optimizer \cite{tieleman2012lecture, mukkamala2017variants} was used with a learning rate of 0.000002. With this setting, the training and the validation data was fed to the model with a batch size of 20. The model was trained for 150 epochs. 
Table \ref{table:summary} gives a brief overview of the different properties of our model.

\begin{table}[h]
\vspace{-5pt}
\caption{Summary of our model} 
\vspace{-5pt}
\label{table:summary}
\renewcommand{\arraystretch}{1.144}
\centering
\begin{tabular}{ | p{2.2cm} | p{5.4cm} | } 
\hline
Properties & What we used \\
\hline
\hline
Transfer Learning Model & VGG16\\
\hline
Weights &  Pre-trained ImageNet weights for the first 14 layers and the rest of the layers were trained using backpropagation\\
\hline
Optimizer & RMSProp with a learning rate of 0.000002\\
\hline
Activation functions & ReLU for the hidden layers, Softmax for the output layer\\
\hline
Regularization & Dropout of 10\% on the layers added on top of VGG16, Data Augmentation using width shift range of 0.3, height shift range of 0.3 and zoom range of 0.2\\
\hline
Input shape & $ 100 \times 200 \times 3 $\\
\hline
Number of output classes & 4\\
\hline
\end{tabular}
\end{table}

\subsection{Results and Evaluation}
\begin{figure}
  \centering
  \includegraphics[clip=true, width=8cm, height=5.5cm]{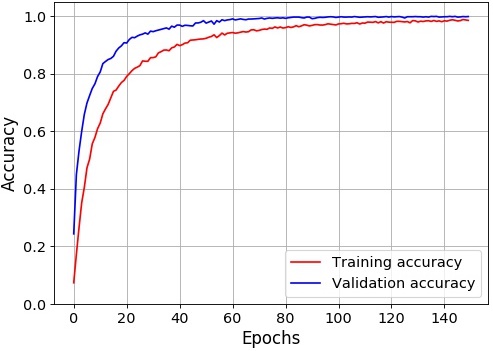}
  \vspace{-5pt}
  \caption{Training and Validation Accuracy}
  \vspace{-5pt}
  \label{fig:acc}
\end{figure}

\begin{figure}
  \centering
  \includegraphics[clip=true, width=8cm, height =5.5cm]{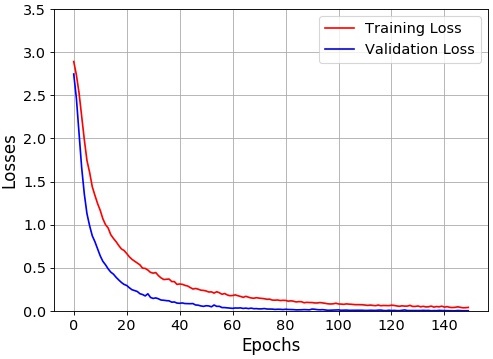}
  \vspace{-5pt}
  \caption{Training and Validation Cross-Entropy Loss}
  \vspace{-5pt}
  \label{fig:loss}
\end{figure}

\begin{figure}[H]
  \centering
  \includegraphics[width=\linewidth]{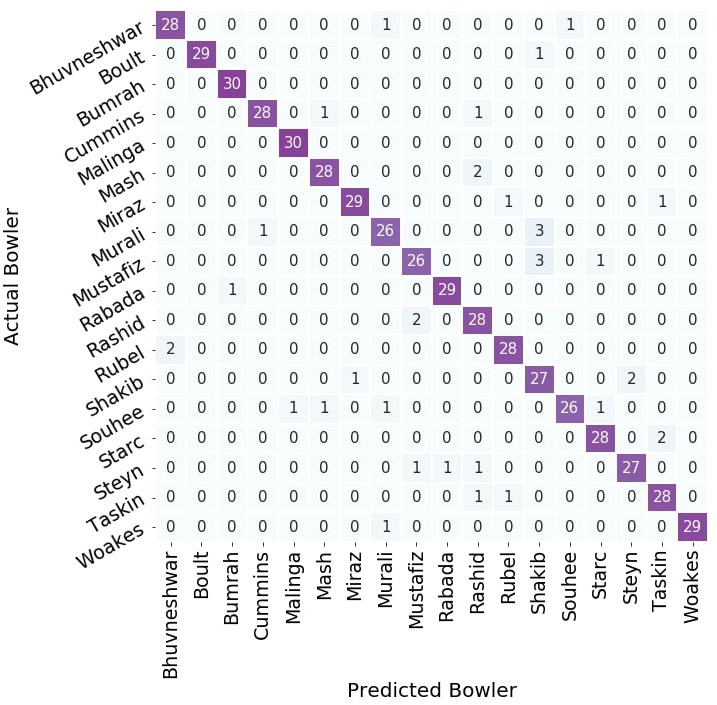}
  \vspace{-10pt}
  \caption{Confusion Matrix}
  \label{fig:confmat}
\end{figure}

After training the model, it achieved an accuracy of almost 100\% on the validation set we used. Fig. \ref{fig:acc} shows the training and validation accuracy increasing steadily with number of epochs. Also, in Fig. \ref{fig:loss} we can see the training and validation loss decreasing to a very low value. To get a more unbiased evaluation of our model, a separate test set containing 540 images (30 images of each class) was prepared. Our model achieved an accuracy of 93.30\% on the test set. Fig. \ref{fig:confmat} shows the confusion matrix of our model when it was tested with the test set. From the confusion matrix, we have measured the precision and recall for each class. Then, the F1 score of the model was calculated using the macro averages of the precision and recall values. A summary of the model evaluation is shown in Table \ref{table:sumeval}. Also to check the strength of the model, we applied several augmentations to the images of the test set and then the images were fed to the model. The model'’s response to different augmentations is summarized in Table \ref{table:eval_aug}.

\begin{table}[H]
\caption{A summary of the model evaluation on the test set}
\label{table:sumeval}
\renewcommand{\arraystretch}{1.3}
\begin{tabular}{|l|c|c|c|c|}
\hline
\multicolumn{1}{|c|}{Name of the class} & Precision & Recall & F1 Score                & Accuracy                \\ \hline \hline
Bhuvneshwar                      & 0.93      & 0.93   & \multirow{18}{*}{0.932} & \multirow{18}{*}{0.933} \\ \cline{1-3}
Boult                            & 1.00      & 0.97   &                         &                         \\ \cline{1-3}
Bumrah                           & 0.97      & 1.00   &                         &                         \\ \cline{1-3}
Cummins                          & 0.97      & 0.93   &                         &                         \\ \cline{1-3}
Malinga                          & 0.97      & 1.00   &                         &                         \\ \cline{1-3}
Mashrafee                        & 0.93      & 0.93   &                         &                         \\ \cline{1-3}
Miraz                            & 0.97      & 0.94   &                         &                         \\ \cline{1-3}
Murali                           & 0.90      & 0.87   &                         &                         \\ \cline{1-3}
Mustafiz                         & 0.90      & 0.87   &                         &                         \\ \cline{1-3}
Rabada                           & 0.97      & 0.97   &                         &                         \\ \cline{1-3}
Rashsid                          & 0.85      & 0.93   &                         &                         \\ \cline{1-3}
Rubel                            & 0.93      & 0.93   &                         &                         \\ \cline{1-3}
Shakib                           & 0.79      & 0.90   &                         &                         \\ \cline{1-3}
Southee                          & 0.96      & 0.87   &                         &                         \\ \cline{1-3}
Starc                            & 0.93      & 0.93   &                         &                         \\ \cline{1-3}
Steyn                            & 0.93      & 0.90   &                         &                         \\ \cline{1-3}
Taskin                           & 0.90      & 0.93   &                         &                         \\ \cline{1-3}
Woakes                           & 1.00      & 0.97   &                         &                         \\ \hline
\end{tabular}
\end{table}


\begin{table}[]
\caption{A summary of the model's response to augmented images}
\label{table:eval_aug}
\renewcommand{\arraystretch}{1.3}
\begin{tabular}{|l|c|c|}
\hline
Augmentation applied                                                              & \begin{tabular}[c]{@{}c@{}}Number of Correct\\  Predictions\\(Out of 540 images)\end{tabular} & \begin{tabular}[c]{@{}c@{}}Accuracy\end{tabular} \\ \hline \hline
\begin{tabular}[c]{@{}l@{}}Rotation \\ (-15 to +15 degrees randomly)\end{tabular} & 487                                                                       & 90.18 \%                                                           \\ \hline
Additive Gaussian Noise                                                           & 501                                                                       & 92.77 \%                                                           \\ \hline
Gaussian Blur                                                                     & 499                                                                       & 92.40 \%                                                           \\ \hline
Perspective Transformation                                                        & 485                                                                       & 89.81 \%                                                           \\ \hline
Cropping                                                                          & 490                                                                       & 90.74 \%                                                           \\ \hline
Sharpening                                                                        & 493                                                                       & 91.29 \%                                                           \\ \hline
\end{tabular}
\end{table}

\vspace{-15pt}
\section{Conclusion}

This paper presents a CNN model which can identify different cricket bowlers based on their bowling actions using transfer learning. We have used VGG16 as our pre-trained transfer learning model. We removed its output layer and added a few dense layers. We have created our own dataset and trained the model with it to identify 18 different cricket bowlers belonging to seven cricket playing nations. Our model has performed remarkably well with a test set accuracy of 93.3\% and F1 score of 93.2\%. In future, we plan to extend our work and include bowlers from all the cricket playing nations.
\vspace{2pt}
\bibliographystyle{./bibliography/IEEEtran}
\bibliography{./bibliography/IEEEabrv}

\end{document}